\documentclass[10pt,twocolumn,letterpaper]{article}

\usepackage{3dv}
\usepackage{times}
\usepackage{epsfig}
\usepackage{graphicx}
\usepackage{amsmath}
\usepackage{amssymb}

\usepackage{cite}
\usepackage{booktabs}
\usepackage{multirow}
\usepackage{subcaption}

\usepackage[pagebackref=true,breaklinks=true,letterpaper=true,colorlinks,bookmarks=false]{hyperref}

\threedvfinalcopy 

\def\threedvPaperID{356} 

\ifthreedvfinal\fi

\begin{document}

\title{Pedestrian Detection in 3D Point Clouds using Deep Neural Networks}


\author{
\`Oscar Lorente$^{1}$ \quad Josep R. Casas$^{1}$ \quad Santiago Royo$^{1,2}$ \quad Ivan Caminal$^{1,2}$ \\ 
$^{1}$ Universitat Polit\`ecnica de Catalunya \quad $^{2}$ Beamagine S.L.
}

\maketitle

\begin{abstract}
    Detecting pedestrians is a crucial task in autonomous driving systems to ensure the safety of drivers and pedestrians. The technologies involved in these algorithms must be precise, robust and reliable, regardless of environment conditions. Relying solely on RGB cameras may not be enough to recognize road environments in situations where cameras cannot capture scenes properly. Some approaches aim to compensate for these limitations by combining RGB cameras with Time Of Flight (TOF) sensors, such as Light Detection and Ranging (LIDAR), which map the scene into 3D point clouds using pulses of light. However, there are few works that address this problem using exclusively the 3D geometric information provided by LIDARs. \\
    In this paper, we propose a PointNet++ based architecture to detect pedestrians in dense 3D point clouds. The aim is to explore the potential contribution of geometric information alone in pedestrian detection systems. PointNet++ is a deep neural network that learns global and local features of a point cloud from its geometry. This allows discerning the different patterns that characterize each class. In our case, the network must learn to classify pedestrians, so we need to carry out supervised training using accurately labeled pedestrian and non-pedestrian point cloud clusters as input. Therefore, we also present a semi-automatic labeling system to generate a point cloud dataset with ground truth. The labeling system transfers pedestrian and non-pedestrian labels from RGB images onto the 3D domain. The fact that our datasets have RGB registered with point clouds enables label transferring by back projection from 2D bounding boxes to point clouds, with only a light manual supervision to validate results. We train PointNet++ with the geometry of the resulting 3D labelled clusters. \\
    The evaluation confirms the effectiveness of the proposed method, yielding precision and recall values around 98\%. This corroborates the hypothesis stated in the paper that 3D geometric information is essential for a neural network to learn to detect  pedestrians in outdoor scenes, thus proving that LIDAR sensors should be a must in these systems.
\end{abstract}

\section{Introduction}
In recent years the interest and research in computer vision and artificial intelligence has grown exponentially, and many related applications have been gradually integrated into our day to day. This is the case of autonomous driving, an innovative technology that combines different sensors such as cameras, radars or GPS, among others, to perceive the surrounding elements and thus be able to act appropriately in each situation. Detecting road environments is a crucial task to ensure the safety of drivers and pedestrians, so the technologies involved in these systems must be precise, robust and reliable, regardless of external factors such as low light conditions. 

Most pedestrian detection systems base their principles on the photometric information of the RGB images captured by the cameras integrated in these vehicles, which plays a fundamental role in object detection. However, relying solely on RGB information can lead to system failures, as this information may not be enough, especially in unfavorable weather conditions. For this reason, some approaches choose to use RGB cameras together with ultrasonic sensors and radars to compensate for these limitations, while others combine the RGB information along with the depth (RGB + D) obtained by Time Of Flight (TOF) sensors~\cite{tof} to improve object detection algorithms~\cite{rgbd1}~\cite{rgbd2}.

A common TOF sensor is the Light Detection and Ranging (LIDAR), which measure the distance between the sensor and its surrounding objects by emitting laser pulses. The information captured by a LIDAR is represented in a 3D point cloud, a data type that provides the 3D geometric information about a scene (X, Y and Z position of each point) regardless of lighting conditions. 

Over the past few years, several pedestrian detection projects have been carried out combining RGB information with that provided by a LIDAR~\cite{lidarrgb1,lidarrgb2, lidarrgb3, lidarrgb4}, but there are few works that address this problem using exclusively the 3D geometric information of point clouds. Therefore, in this paper we present a Deep Neural Network (DNN) based architecture to detect pedestrians in dense 3D point clouds obtained with a LIDAR. 

In order to explore the potential contribution of geometric information alone, we design and implement a pedestrian detection system using PointNet++~\cite{pointnet++}, a pioneering DNN in point cloud classification and segmentation that learns global and local features of a point cloud from its geometry. This allows discerning the different patterns that characterize each class and being able to predict them accordingly. 

In our case, PointNet++ must learn to classify pedestrians from non-pedestrians, so we must carry out supervised training using accurately labeled pedestrian and non-pedestrian clusters (ground truth) as input. A classic approach would be to manually label, but considering that thousands of input samples are needed to train a neural network, this would take too much time and work. The state of the art in detection in RGB images is very mature, so a better strategy would be to take advantage of this maturity to generate a ground truth in point clouds. In this way, we present a semi-automatic 3D labeling system based on the transfer of pedestrian and non-pedestrian 2D bounding boxes from the RGB images onto the 3D domain with only a light manual supervision to validate results.

\section{Related Work}

RGB information is a key element in understanding the world around us. Image-based techniques are the most popular in pedestrian detection problems. They show good overall performance, but in situations where the images are not clear the detections are unreliable. For this reason, many researchers combine video cameras with LIDARs~\cite{lidarrgb1,lidarrgb2, lidarrgb3, lidarrgb4} to address these problems. However, there are few LIDAR-based approaches that rely solely on the 3D geometric information of point clouds. 

Navarro-Serment \etal~\cite{navarro-serment} present a pedestrian detection and tracking method based on motion and geometric features. To reduce the computational load of processing the entire point cloud, they first project data subsets onto the 2D plane to perform object detection. Then, the detections are used to compute a motion score and a set of geometric features in the corresponding 3D point clouds. Therefore, the algorithm relies on the detections performed in the 2D domain. This implies projecting the original point cloud onto a 2D plane, which can lead to a loss of 3D geometric information.

Kidono \etal~\cite{kidono} propose two new features to improve~\cite{navarro-serment} performance in long ranges. Both the slice feature and the distribution of reflection intensities are extracted from pedestrian candidates clusters. To obtain these clusters, the 3D point cloud is projected onto a 2D occupancy grid to divide it into two classes: ground plane and objects. The cell size of the occupancy grid and the threshold of the height for segmentation are fixed values, and they do not adapt to the 3D geometric shape of pedestrians in different scenarios.

Tang \etal~\cite{tang} and Lin \etal~\cite{lin} approaches are similar, as they both train a Support Vector Machine (SVM) with extracted 2D and 3D features. However, Tang \etal features are all hand-crafted, while Lin \etal design a Convolutional Neural Network (CNN) to extract 3D features. Although the latter allows obtaining relevant deep features, the 3D point cloud data must be discretized into a 3D voxel representation to train the CNN. 

We present a pedestrian detector in 3D point clouds based on PointNet++, a hierarchical neural network that extracts deep features from the 3D geometric information of a point cloud. The local features are extracted capturing fine geometric structures from small neighborhoods. Such local features are further grouped into larger units and processed to produce higher level features. This process is repeated until the local and global features of the whole point set are obtained. In this way, the 3D geometric information is respected during the process.

\section{Data}

\begin{figure}
\begin{center}
\begin{subfigure}{0.4\linewidth}
\begin{center}
\includegraphics[width=\linewidth]{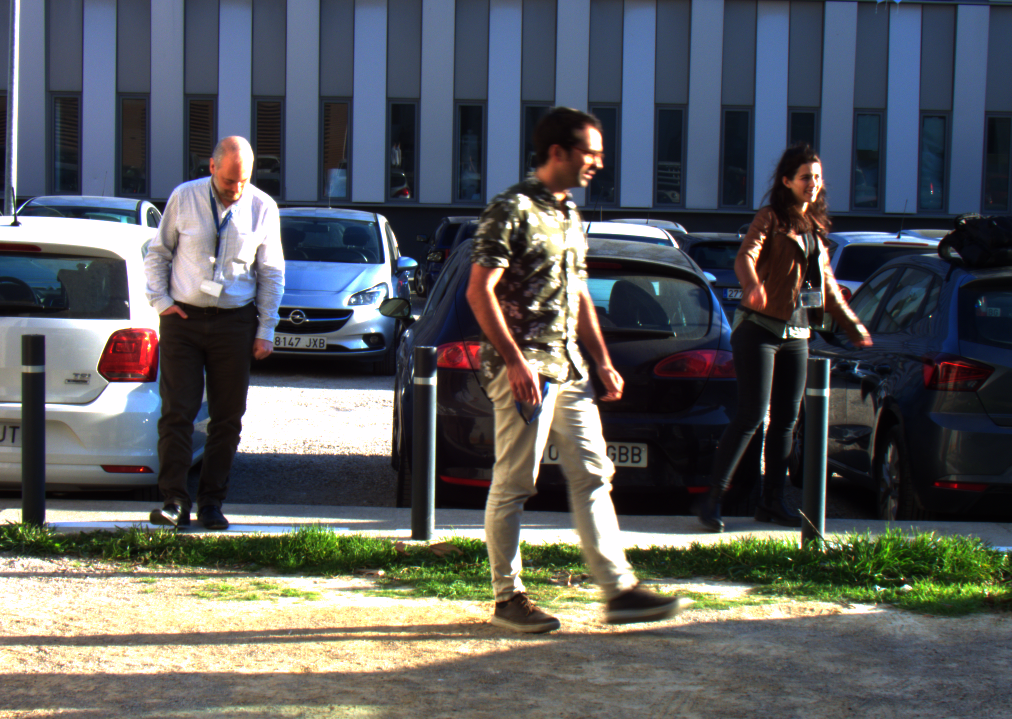}
\end{center}
\end{subfigure}
\begin{subfigure}{0.4\linewidth}
\begin{center}
\includegraphics[width=\linewidth]{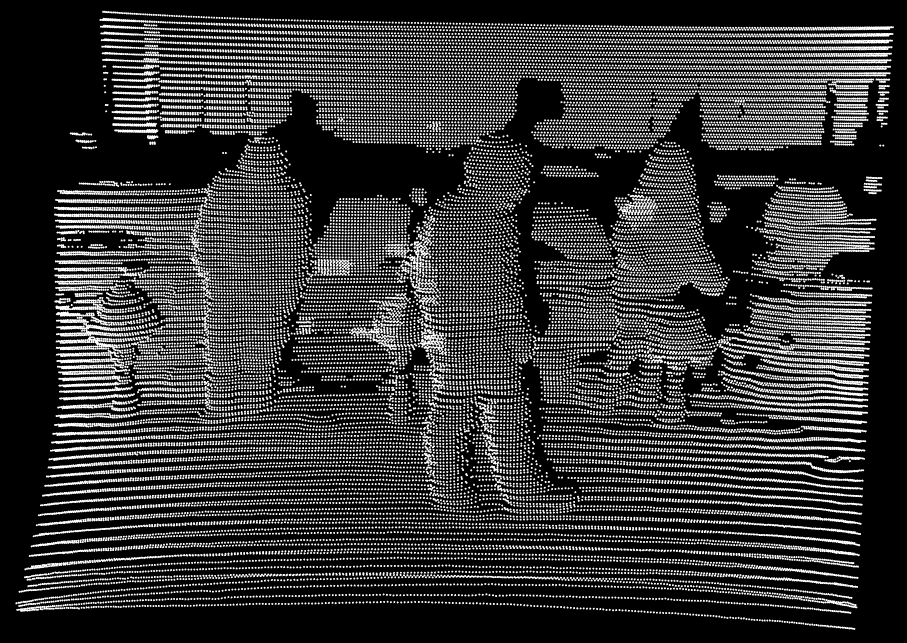}
\end{center}
\end{subfigure}
\end{center}
\caption{Sample RGB image (left) and 3D point cloud (right) obtained with Beamagine's L3CAM sensor. Note the vertical resolution of the solid state LIDAR.}
\label{fig:sample-data}
\end{figure}

\begin{figure*}[t]
\begin{center}
\includegraphics[width=0.9\linewidth]{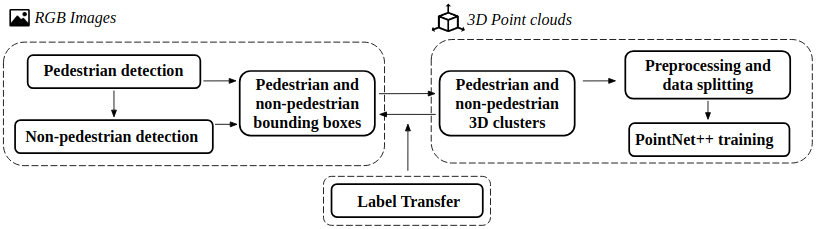}
\end{center}
   \caption{Overview of the proposed pedestrian detector method.}
\label{fig:workflow}
\end{figure*}

In this paper we use the L3CAM sensor from Beamagine S.L. to obtain the datasets. This sensor incorporates a video camera and a LIDAR that provide RGB images and dense 3D point clouds registered and synchronized (Figure~\ref{fig:sample-data}). The sensors are calibrated according to García \etal~\cite{beamagine}. This enables performing the transfer of labels using back projection from 2D bounding boxes to 3D clusters. The density and, in particular, the vertical resolution of the L3CAM LIDAR, has proven to be a fundamental asset for our contribution.

The RGB images have a resolution of 1224 x 1024 pixels. The LIDAR emits 300 x 150 pulses of light with a Field Of View (FOV) of 30º x 20º, which translates into an angular resolution of approximately 0.1º x 0.13º (without taking into account distortion). Consequently, each point cloud has a maximum of 45 thousand points, but this value varies depending on the elements contained in the scene and whether they have been captured properly by the LIDAR.

The obtained dataset consists of 10 scenes captured outdoors in different streets of Terrassa, Spain. Specifically, there are 3,010 images and point clouds captured at 1 frame per second (fps). We attach more information about Beamagine datasets (images, point clouds and calibration parameters) as supplementary material.

\section{Proposed Method}

Detecting pedestrians in 3D point clouds with a DNN requires labeled pedestrian and non-pedestrian clusters so that we can train PoinNet++ in a supervised manner. To generate this ground truth we propose a semi-automatic labeling strategy in which pedestrian bounding boxes (PBB) are first predicted in RGB images using a pedestrian detector, then non-pedestrian bounding boxes (NPBB) are also generated in RGB images and finally these labels are transferred to the corresponding points clouds. The overview of the proposed method is presented in Figure~\ref{fig:workflow}.

\subsection{Pedestrian Detection in RGB Images}

We propose using You Only Look Once (YOLOv3)~\cite{yolov3} to label pedestrians with 2D bounding boxes (PBB) in the RGB images of our datasets. We must ensure that YOLO predictions are accurate and reliable, as they will be used later in the training of PointNet++. The performance of YOLO for label generation is evaluated comparing 10\% of the network predictions with a manually labeled pedestrian ground truth. 

\begin{figure}[t]
\begin{center}
\begin{subfigure}{0.3\linewidth}
\begin{center}
\includegraphics[width=\linewidth]{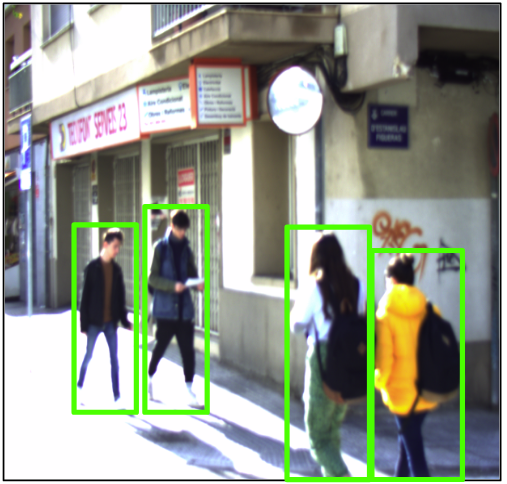}
\end{center}
\caption{}
\end{subfigure}
\begin{subfigure}{0.3\linewidth}
\begin{center}
\includegraphics[width=\linewidth]{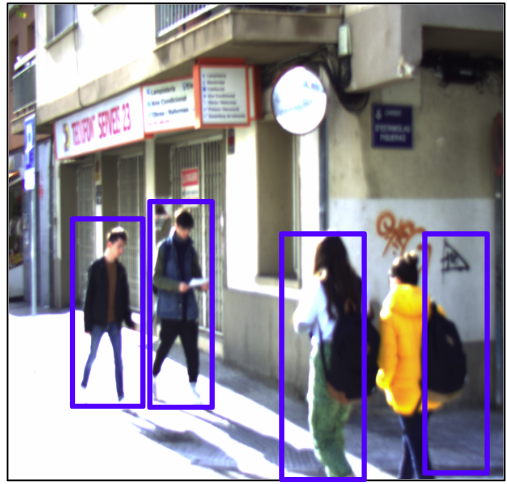}
\end{center}
\caption{}
\end{subfigure}
\begin{subfigure}{0.3\linewidth}
\begin{center}
\includegraphics[width=\linewidth]{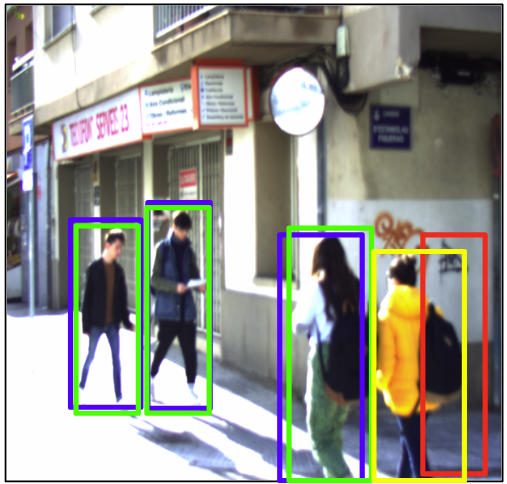}
\end{center}
\caption{}
\end{subfigure}
\end{center}
\caption{Evaluation of YOLO pedestrian detection in our RGB images: (a) Manually labeled pedestrians. (b) YOLO predictions. (c) Detections with an IOU above 0.5 are considered true positives (green/blue), and those with an IOU below 0.5 are false positives (red). Manually labeled pedestrians not detected by YOLO are false negatives (yellow).}
\label{fig:yolo-evaluation}
\end{figure}

The metrics used in this evaluation is the Intersection Over Union (IOU), which measures the level of overlap between a YOLO-generated bounding box and a manually labeled one. As shown in Figure~\ref{fig:yolo-evaluation}, if this overlap is greater than 0.5, we consider the detection to be correct (true positive). If a YOLO prediction does not correspond to any bounding box in the ground truth, we have a false positive, and a false negative is a manually labeled pedestrian that has not been detected by YOLO.

Following this criterion, we evaluate YOLO performance for label generation using precision (proportion of correct detections) and recall (proportion of detected pedestrians).

In order for PointNet++ to properly learn to differentiate between what is a pedestrian and what is not, the labeled 3D clusters must be highly accurate. There should not be many non-pedestrians mislabeled as pedestrians. Therefore, in the evaluation of YOLO, precision must be prioritized over recall. We should ensure that even if YOLO does not detect all the pedestrians in the scenes and there are some miss detections (lower recall), the detections it makes are correct and there are very few false positives (highest precision).

For this reason, we adjust the network parameters: confidence and non-maximum suppression thresholds (to 0.8 and 0.3, respectively), and we obtain a precision of 99.8\% while keeping a recall of 77.9\%.

\subsection{Non-Pedestrian Detection in RGB images}
\label{sec:bbnp}

\begin{figure}[t]
\begin{center}
\begin{subfigure}{0.35\linewidth}
\begin{center}
\includegraphics[width=\linewidth]{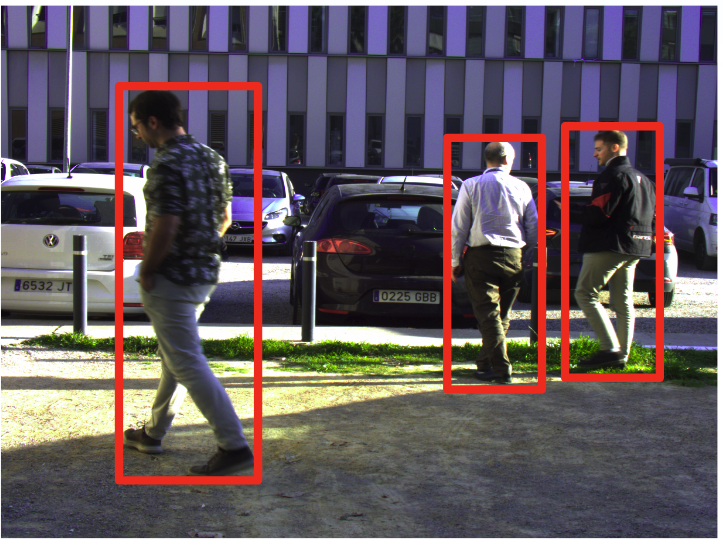}
\end{center}
\end{subfigure}
\begin{subfigure}{0.35\linewidth}
\begin{center}
\includegraphics[width=\linewidth]{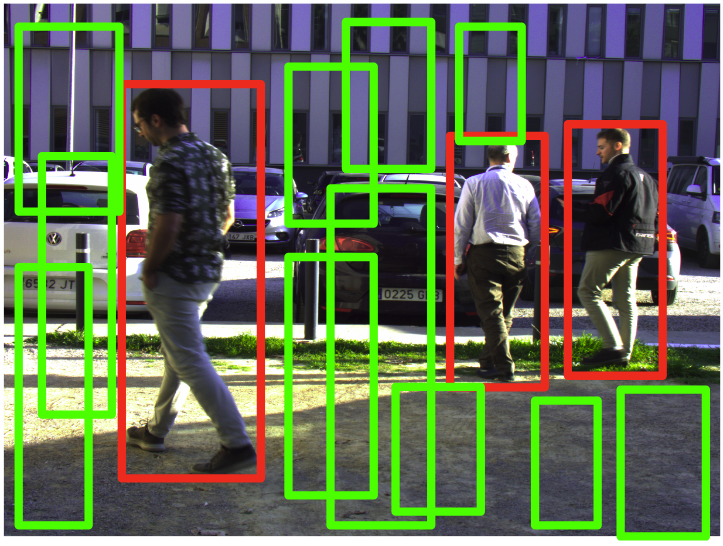}
\end{center}
\end{subfigure}
\end{center}
\caption{Pedestrian (left, red) and non-pedestrian (right, green) 2D bounding boxes.}
\label{fig:bbnp}
\end{figure}

We cannot train PointNet++ with only pedestrian point cloud clusters, we also need examples of the negative class: non-pedestrians. For this reason, we design and implement an algorithm to generate non-pedestrian 2D bounding boxes (NPBB) for the RGB images in our dataset (Figure~\ref{fig:bbnp}) that will also be transferred to the 3D domain to get the non-pedestrian clusters. 

To train a neural network rigorously, the training data must represent the world as it is. The PBB detected by YOLO occupy 11.5\% of the pixels of the entire dataset, so one way to respect this reality would be to generate a total of non-pedestrian labels such that the ratio of pedestrians to non-pedestrians remains the same: 11.5. Thus, for 13,232 pedestrian labels, this implies generating a total of 102,130 NPBB.

On the other hand, we want PointNet++ to learn to detect pedestrians by the geometric information of the point clouds and not by the size or shape of the clusters, so the 2D bounding boxes must be similar in terms of width, height and aspect ratio. For this reason, we design a strategy to randomly generate NPBB that mimic the statistics of PBB, as shown in Table~\ref{tab:bb-stats}.

\begin{table}
\begin{center}
\begin{tabular}{ccccc}
\toprule
\textbf{Class} & 
\textbf{} & 
\textbf{Width} & 
\textbf{Height} & 
\textbf{AR} \\ 
\midrule\midrule
\multirow{2}{*}{Pedestrian} &
Mean & 
$95.3$ & 
$266.3$ & 
$2.9$   \\ & 
SD & 
$56.9$ & 
$145.6$ & 
$0.9$ \\
\midrule
\multirow{2}{*}{Non-pedestrian} & 
Mean & 
$111.1$ & 
$302.5$ & 
$2.8$ \\ & 
SD & 
$37.9$ & 
$102.3$ & 
$0.9$ \\ 
\bottomrule
\end{tabular}
\end{center}
\caption{Mean and standard deviation (SD) of width, height and aspect ratio (AR) of the pedestrian (PBB) and non-pedestrian (NPBB) 2D bounding boxes.}
\label{tab:bb-stats}
\end{table}

Generating 102,130 bounding boxes of non-pedestrians in only 3010 images implies having approximately 34 NPBB per image. There are 4 PBB per image, and the NPBB must respect the statistics of PBB, so we have a high level of overlap among NPBB: 70\% on average. If it were necessary to use NPBB with less overlap, we could not generate as many bounding boxes as the ones initially proposed. The implications of this are presented in the results section.

\subsection{2D to 3D Label Transfer}

\begin{figure}[t]
\begin{center}
\begin{subfigure}{0.4\linewidth}
\begin{center}
\includegraphics[width=\linewidth]{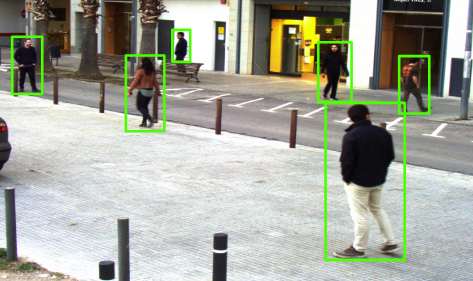}
\end{center}
\caption{}
\label{fig:transfer-rgb}
\end{subfigure}
\begin{subfigure}{0.4\linewidth}
\begin{center}
\includegraphics[width=\linewidth]{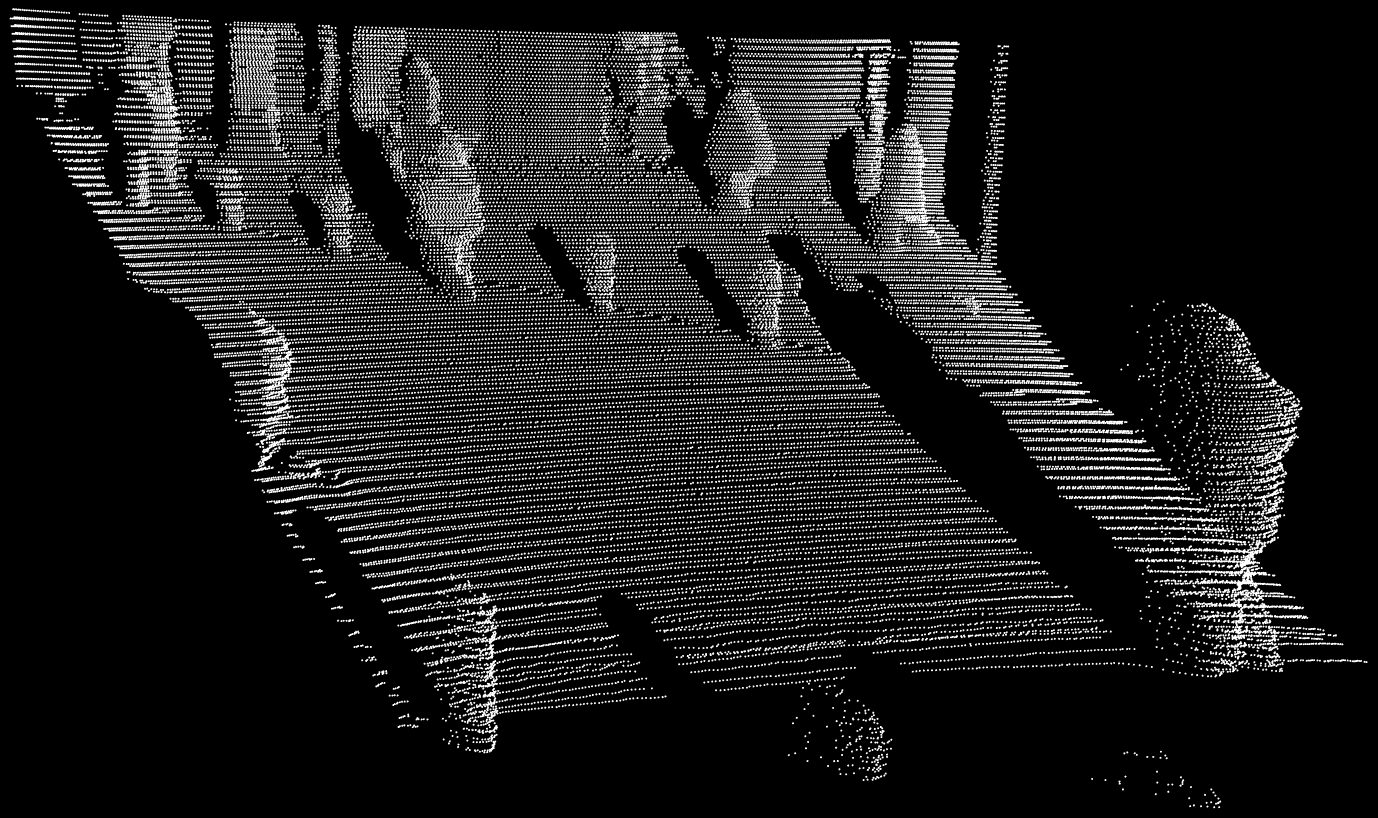}
\end{center}
\caption{}
\label{fig:transfer-pc}
\end{subfigure}
\begin{subfigure}{0.4\linewidth}
\begin{center}
\includegraphics[width=\linewidth]{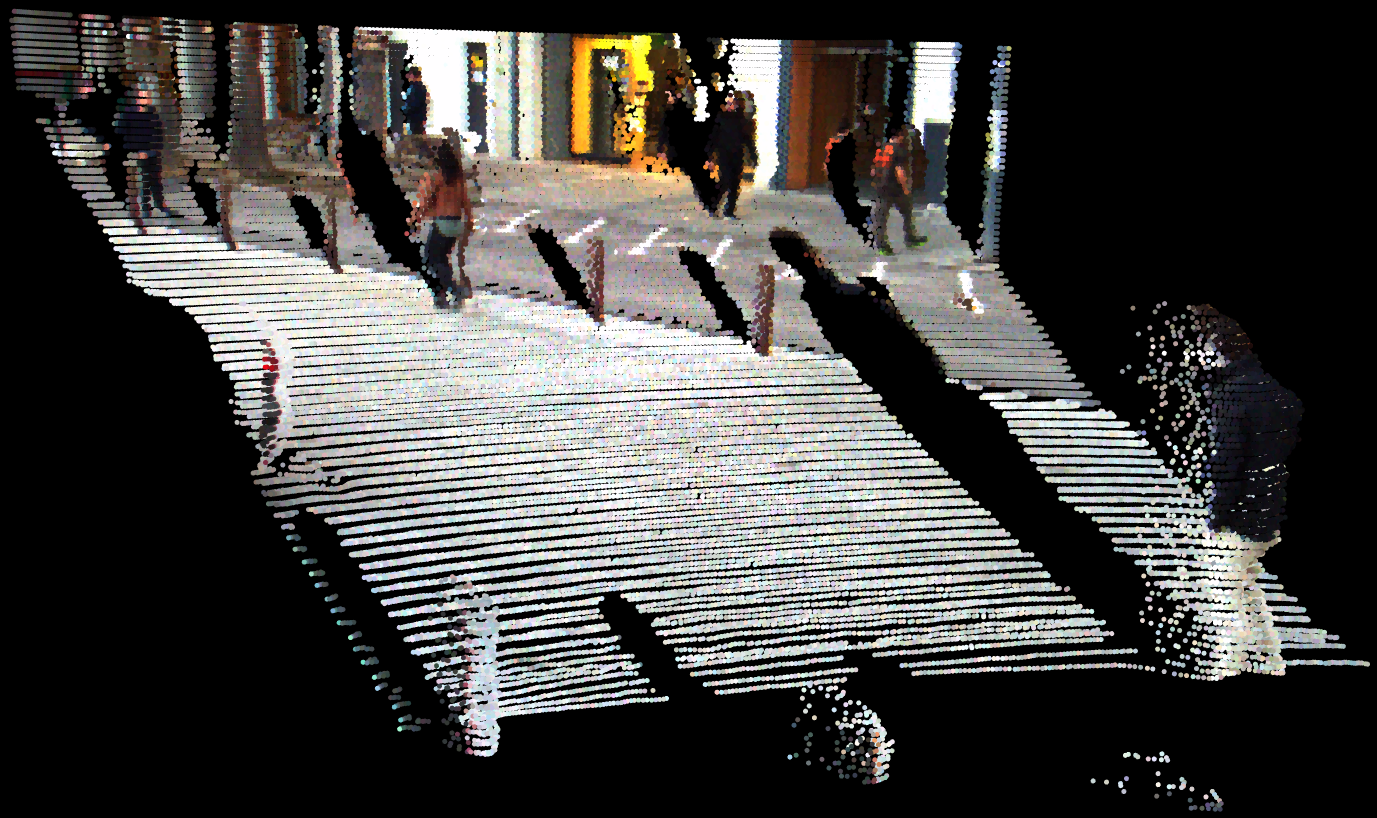}
\end{center}
\caption{}
\label{fig:transfer-projection}
\end{subfigure}
\begin{subfigure}{0.4\linewidth}
\begin{center}
\includegraphics[width=\linewidth]{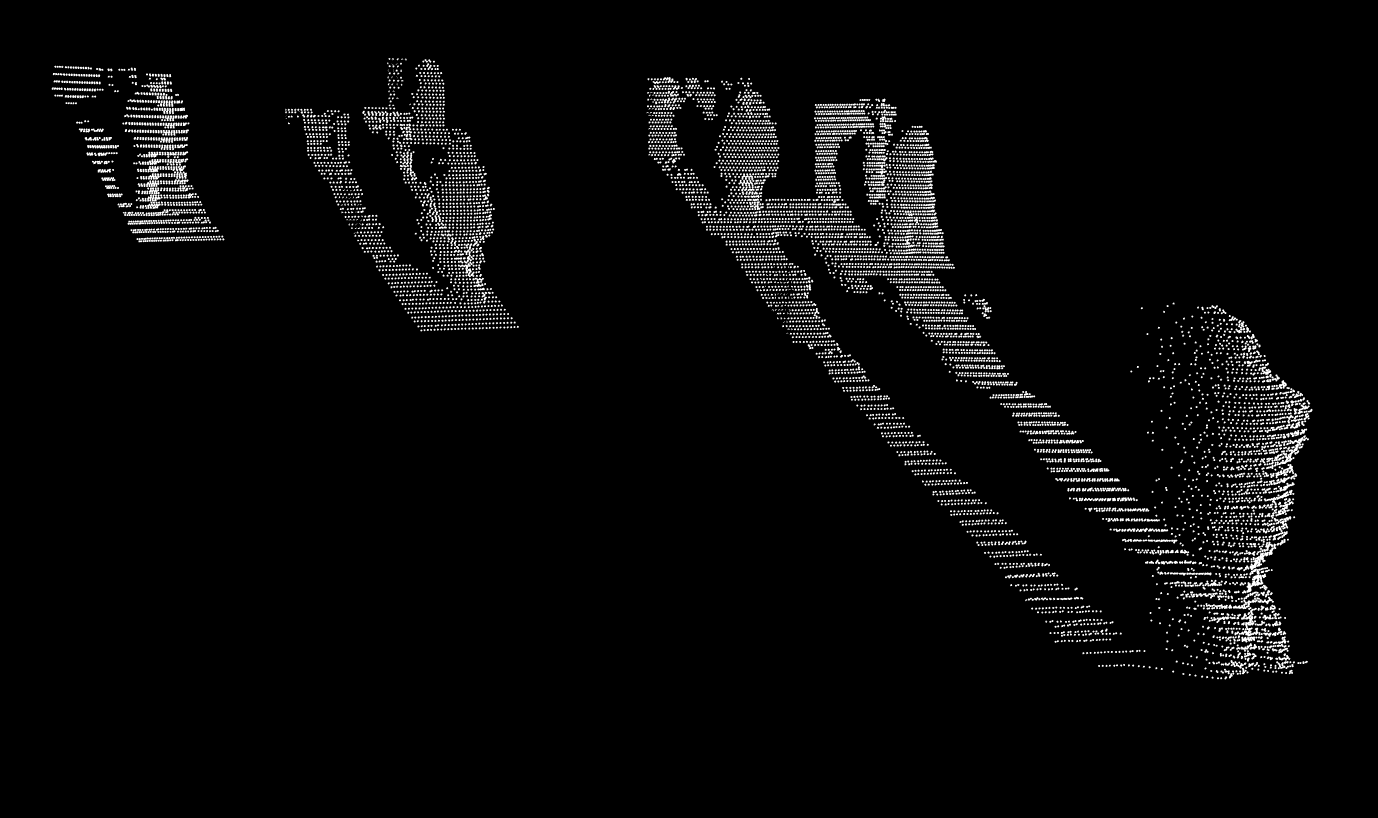}
\end{center}
\caption{}
\label{fig:transfer-clusters}
\end{subfigure}
\end{center}
\caption{Label transfer pipeline: (a) Pedestrians detected by YOLO in a RGB image. (b) Corresponding 3D point cloud. (c) Point cloud projected onto the RGB image. (d) Transferred pedestrian 3D clusters.}
\label{fig:label-transfer}
\end{figure}

The transfer of labels from RGB images (Figure~\ref{fig:transfer-rgb}) into the corresponding point clouds (Figure~\ref{fig:transfer-pc}) is carried out using projection matrices calibrated for the L3CAM sensor~\cite{beamagine}. These matrices allow projecting the point clouds onto the images to know which pixel each cloud point corresponds to (Figure~\ref{fig:transfer-projection}). If this pixel is inside any bounding box, the analogous point of the point cloud is labeled with the corresponding label (back projection). By repeating this process for all the images of the dataset we obtain the pedestrian and non-pedestrian clusters with which to train PointNet++ (Figure~\ref{fig:transfer-clusters}). 

To accept a pedestrian or non-pedestrian cluster as valid, a restriction must be taken into account. PointNet++ design implies that all point clouds used as input to the DNN must have the same number of points, and this number must be a power of two. For this reason, we train PointNet++ with point clouds of 1024 points each. Interpolating point clouds that have fewer than 1024 points to generate the required 1024 could adversely affect the shape of a pedestrian. Consequently, clusters with no more than 1024 points are discarded. Defining a minimum of 1024 points per cluster also ensures that they have enough representative information for the network to be able to learn geometries from them. 

\subsection{Preprocessing of 3D Clusters}

Before training PointNet++ we must prepare the pedestrian and non-pedestrian clusters so that the neural network can learn to detect pedestrians adequately. This preprocessing involves two phases: downsampling and normalization.

\subsubsection{Downsampling}

\begin{figure}[t]
\begin{center}
\begin{subfigure}{0.18\linewidth}
\begin{center}
\includegraphics[width=\linewidth]{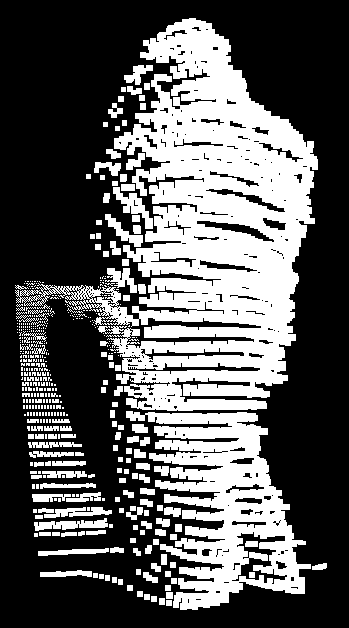}
\end{center}
\caption{}
\end{subfigure}
\begin{subfigure}{0.18\linewidth}
\begin{center}
\includegraphics[width=\linewidth]{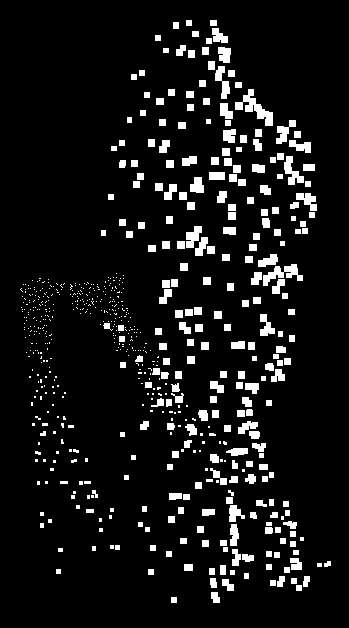}
\end{center}
\caption{}
\label{fig:rnd}
\end{subfigure}
\begin{subfigure}{0.18\linewidth}
\begin{center}
\includegraphics[width=\linewidth]{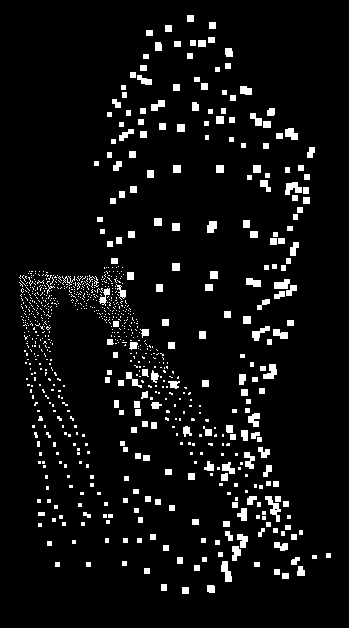}
\end{center}
\caption{}
\label{fig:voxel}
\end{subfigure}
\begin{subfigure}{0.18\linewidth}
\begin{center}
\includegraphics[width=\linewidth]{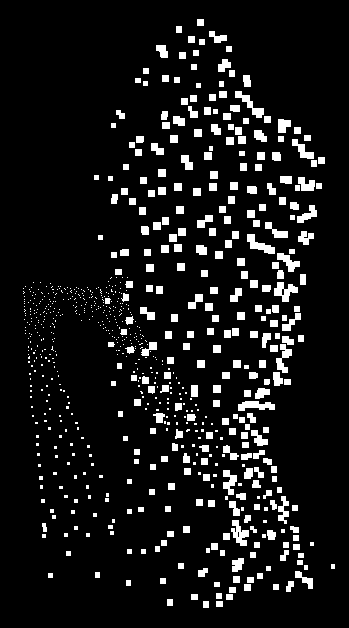}
\end{center}
\caption{}
\label{fig:fps}
\end{subfigure}
\end{center}
\caption{Comparison between the downsampling techniques. (a) Original: 7,063 points. (b) Random sampling: 1,024 points. (c) Voxel grid filter: 1,578 points. (d) Farthest Point Sampling: 1,024 points.}
\label{fig:downsampling}
\end{figure}

Each cluster has a different number of points (greater than 1024), so the first step in preprocessing the input data is to downsample each point cloud cluster to a fixed number of points.

The downsampling technique must be rigorous and respect the geometric shape of the point cloud on which it is applied (essential in the case of pedestrians). We consider the following downsampling algorithms:

\begin{itemize}
	\item \textbf{Random Sampling:} Randomly chooses a fixed number of points from the cluster. Although generating a subset of a fixed number of points, it has the drawback that this subset is not representative of the original cluster, since when choosing the points randomly, the geometric shape of the object is not respected (Figure~\ref{fig:rnd}).
	
	\item \textbf{Voxel Grid:} Divides the point cloud into a 3D grid and gets from each voxel the representative point (centroid) of the points that are inside the voxel. The voxel grid filter certainly respects the geometric shape of the cluster, so we obtain a more representative subset of the original point cloud than in the case of random sampling (Figure~\ref{fig:voxel}). However, the resulting point cloud does not have a fixed number of points, since we do not know a priori how many points will fall within each voxel. There are techniques to compensate for this problem (\eg dynamic voxel size), but we prefer to use a simpler algorithm that also provides better results.
	
	\item \textbf{Farthest Point Sampling (FPS):} Provides a subset of a fixed number of points such that the minimum pairwise distances for each point are maximized~\cite{fps}. The Farthest Point Sampling technique generates a subset that is completely representative of the original point cloud, since the distribution of the points fully respects the geometric shape of the original cluster (Figure~\ref{fig:fps}). For this reason, we use the FPS algorithm to downsample the 3D clusters.
	
\end{itemize}

\subsubsection{Normalization}

Due to the design of the PointNet++ architecture, the input point clouds must be located at the coordinate origin with all the points within a unit sphere. Therefore, we center each cluster at the coordinate origin by subtracting the point cloud mean (centroid) from each point. Then, we normalize it to a unit sphere by dividing the coordinates of each point by the Euclidean distance from the center to the farthest point.

\subsection{Impact of Preprocessing and Imbalance in PointNet++ with Modelnet40}

Qi \etal~\cite{pointnet++} present their results using the Modelnet40 dataset. This dataset consists of CAD models from 40 different categories, and has a total of 12,308 objects: 9,840 for training and 2,468 for validation (and testing). As a result, we carry out some previous experiments using Modelnet40 as the dataset, to check if the differences that exist between this dataset and ours have a negative impact on the training of the network. The results are presented in Table~\ref{tab:modelnet-results}.

\subsubsection{Our preprocessing method}

\begin{figure}[t]
\begin{center}
\begin{subfigure}{0.3\linewidth}
\begin{center}
\includegraphics[width=\linewidth]{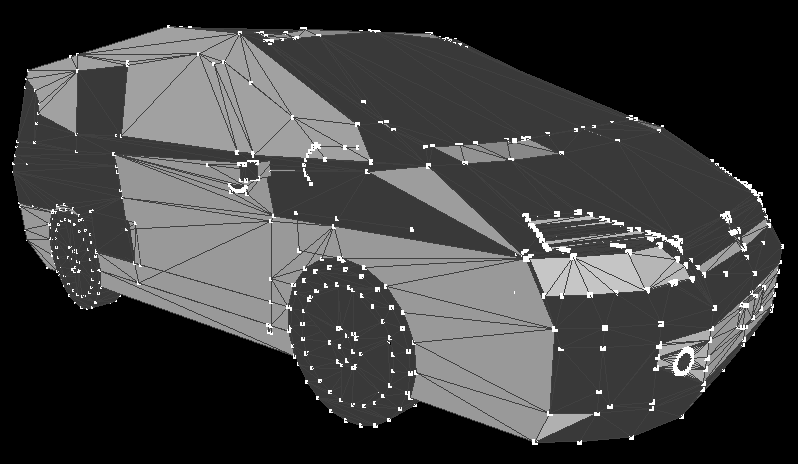}
\end{center}
\caption{}
\end{subfigure}
\begin{subfigure}{0.3\linewidth}
\begin{center}
\includegraphics[width=\linewidth]{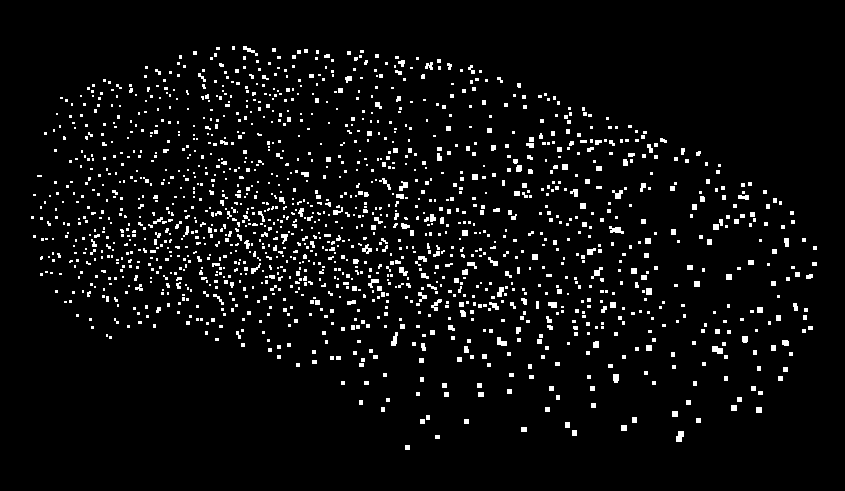}
\end{center}
\caption{}
\label{fig:modelnet-mesh}
\end{subfigure}
\begin{subfigure}{0.3\linewidth}
\begin{center}
\includegraphics[width=\linewidth]{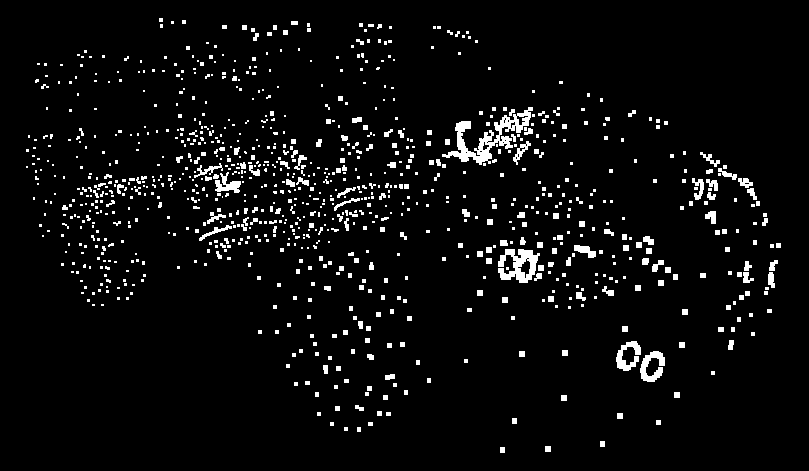}
\end{center}
\caption{}
\label{fig:modelnet-preprocessed}
\end{subfigure}
\end{center}
\caption{Modelnet40 car sample: (a) Original - mesh faces. (b) Qi \etal preprocessing - points sampled on mesh faces. (c) Our preprocessing - subset of mesh vertices.}
\end{figure}

The authors of PointNet++ perform a specific preprocessing to the ModelNet40 CAD models in order to train the neural network with them. Specifically, they first generate 10,000 points for each object by sampling points on the mesh faces, and then apply the Farthest Point Sampling algorithm to downsample each point cloud to a fixed number of points: 1024. Finally, they apply mean subtraction and unit sphere normalization. However, we cannot follow the same procedure since our point clouds are captured using a LIDAR, and therefore do not have mesh faces. Thus, our preprocessing is based on applying FPS on the point clouds (without doing any previous sampling) to obtain a fixed number of points and then subtracting the mean and normalizing.

Comparing Figures~\ref{fig:modelnet-mesh} and~\ref{fig:modelnet-preprocessed} we can see the difference between the two preprocessing methods. As observed, this disparity is significant. While Qi \etal carry out a first mesh sampling in which the 10,000 points chosen are distributed uniformly throughout the object, we start from only the point cloud of the object, and select a subset of these vertices.

\begin{table}
\begin{center}
\begin{tabular}{lcc}
\toprule
\textbf{Experiment} & 
\textbf{Accuracy} & 
\textbf{Avg. class accuracy} \\ 
\midrule\midrule
Original & 
88.9 & 
86.6 \\
\midrule
Our preprocessing & 
74.1 & 
62.6 \\
\midrule
Binary dataset & 
99.4 & 
98.1 \\
\bottomrule
\end{tabular}
\end{center}
\caption{Results obtained in the Modelnet40 experiments.}
\label{tab:modelnet-results}
\end{table}

The results worsen in our case due to the less homogeneous and therefore less representative distribution of points within the point cloud.  However, this worsening is within expectations, since our preprocessing techniques are designed for LIDAR data and not for CAD models used by PointNet++ authors. Furthermore, in our case the class of the object is still recognizable, so we can affirm that our preprocessing method is correctly implemented.

\subsubsection{Binary Classification - Imbalanced Dataset}

Modelnet40, the dataset used by Qi \etal to present the results of PointNet++, is a balanced dataset consisting of 40 different classes, with a relatively uniformly distributed number of objects per class.

On the other hand, our dataset is a binary dataset (pedestrian and non-pedestrian) and the ratio of pedestrian to non-pedestrian clusters is approximately 10/90, so we have an imbalanced dataset. This may not be a problem as long as PointNet++ design is prepared to deal with imbalanced classification. However, we want to make sure that the ratio between point clouds of the positive and negative class does not impair the learning of the neural network.

For these reasons, the last experiment we perform using the dataset Modelnet40 is to rearrange its point clouds into a binary structure. We merge the Modelnet40 \textit{airplane}, \textit{bathtub} and \textit{bed} classes into a new class: positive,  while the other 37 classes form the negative class. In this way, we convert the problem into a binary classification, where the number of point clouds of the positive and negative classes are 1,496 and 10,812, respectively, maintaining a proportion similar to that of our dataset. 

This experiment aims to verify that PointNet++ works correctly with a binary and imbalanced dataset. The accuracy, precision (98.0) and recall (96.4) of the system demonstrate that training PointNet++ with a dataset of these characteristics does not impair the learning of the network.



\section{Results and Experiments}

\subsection{Dataset splitting}

\begin{table}
\begin{center}
\begin{tabular}{lccc}
\toprule
\textbf{Dataset} & 
\textbf{Pedestrian} & 
\textbf{Non-pedestrian} & 
\textbf{Total} \\
\midrule\midrule
Training & 
6,932 & 
60,388 & 
67,320 \\
\midrule
Validation & 
1,733 & 
15,098 & 
16,831 \\
\midrule
Test & 
345 & 
3,040 & 
3,385 \\
\midrule
\midrule
\textbf{Total} & 
\textbf{9,010} & 
\textbf{78,526} & 
\textbf{87,536} \\
\bottomrule
\end{tabular}
\end{center}
\caption{Number of pedestrian and non-pedestrian 3D clusters in the training, validation and test datasets.}
\label{tab:datasets}
\end{table}

We organize the pedestrian and non-pedestrian clusters into training, validation and test, as shown in Table~\ref{tab:datasets}. In the PointNet++ training stage, the neural network uses the training dataset to learn to discriminate pedestrians and non-pedestrians. Then it evaluates how well it has learned using the validation dataset, and iteratively modifies internal parameters of the network to improve its performance. Finally, once the training stage has finished, we test the performance of PointNet++ with a dataset that the network has not seen while training, the test dataset, and from the results obtained we can assess how well it has learned to classify and generalize this learning.

The darkest scene is the most difficult one for YOLO to detect pedestrians, since it has a 48\% of recall compared to the average recall: 78\%. We decided to use this scene as the test dataset to see if PointNet++ can learn to classify in these situations and contribute to a more reliable and secure detection.

The other 9 scenes are used for the training and validation datasets following a ratio of 80/20, respectively.

To structure the datasets in a meticulous way, we take advantage of the fact that we have manually labeled 10\% of the RGB images to evaluate YOLO. This proportion is respected in the training, validation and test datasets. In other words, in each of the three datasets, 10\% of the pedestrian clusters correspond to point clouds obtained from the transfer of the manually labeled 2D pedestrian bounding boxes.

\subsection{Experiments with Beamagine datasets}

The purpose of these experiments is to study under which conditions PointNet++ provides better results detecting pedestrians in outdoor datasets. To this end, the results are presented in Table~\ref{tab:outdoor-experiments} in terms of precision, recall and F1-score.

\subsubsection{Batch size}

Our dataset consists of 87,536 point clouds, while Modelnet40 has 12,308. This could imply that some network parameters such as the batch size, which defines the number of samples that the neural network works with before updating its internal parameters (weights), had to be adapted to deal with this difference.

Qi \etal use a batch size of 16 or 32 with Modelnet40. Consequently, in the first test we perform with our dataset we define a batch size of 16. As this test yields poor results in terms of recall, we evaluate PointNet++ performance for batch sizes ranging from 16 to 128. That way we see if by increasing the number of point clouds, we should also increase the batch size, or if these unsatisfactory results are due to another reason.

Table~\ref{tab:outdoor-experiments} shows that the higher precision value is obtained using a batch size of 16. However, the recall and F1-score are lower than in the 32 case, so we infer that a batch size of 32 is adequate for PointNet++ to learn from our data. Henceforth, we use a batch size of 32 for the other tests, and this first experiment is referred to as the \textbf{baseline}.

\begin{table}
\begin{center}
\begin{tabular}{lccc}
\toprule
\textbf{Experiment} & 
\textbf{Precision} & 
\textbf{Recall} &
\textbf{F1} \\ 
\midrule\midrule
Batch size: 16 & 
99.2 & 
38.8 &
55.8 \\
\midrule
Batch size: 32 & 
91.7 & 
41.4 &
57.1 \\
\midrule
Batch size: 64 & 
39.3 & 
47.8 &
43.1 \\
\midrule
Batch size: 128 &
32.7 &
48.4 &
39.1 \\
\midrule\midrule
No augmentation & 
99.1 & 
98.6 &
98.8 \\
\midrule\midrule
20\% NPBB overlap & 
97.1 & 
97.7 &
97.4 \\
\midrule
MSG model & 
99.4 & 
92.5 &
95.8 \\
\bottomrule
\end{tabular}
\end{center}
\caption{Results obtained in the experiments performed in our datasets.}
\label{tab:outdoor-experiments}
\end{table}

\subsubsection{Data augmentation}


\begin{figure}[t]
\begin{center}
\begin{subfigure}{0.9\linewidth}
\begin{center}
\includegraphics[width=\linewidth]{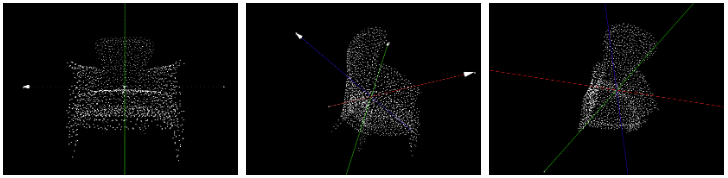}
\end{center}
\caption{}
\label{fig:augmentation-modelnet}
\end{subfigure}
\begin{subfigure}{0.9\linewidth}
\begin{center}
\includegraphics[width=\linewidth]{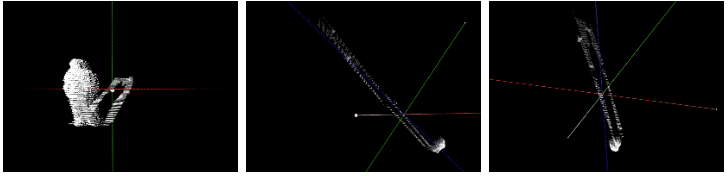}
\end{center}
\caption{}
\label{fig:augmentation-outdoor}
\end{subfigure}
\end{center}
\caption{Data augmentation in terms of rotation (three different angles) applied on: (a) Modelnet40 chair. (b) Beamagine pedestrian cluster.}
\end{figure}

Qi \etal apply data augmentation (rotations, jitter) in the training dataset so that PointNet++ learns to better differentiate each object class regardless of whether it presents a certain rotation or disturbance.

The data augmentation is convenient and useful in datasets like Modelnet40, in which the point clouds obtained from CAD models have little noise and maintain its appearance and shape regardless of the rotation applied (Figure~\ref{fig:augmentation-modelnet}). However, our data has been obtained using a LIDAR sensor, which is noisier and only captures the points not occluded from the sensor point of view. Therefore, a pedestrian cluster completely loses the shape and appearance of a person when the point cloud is rotated (Figure~\ref{fig:augmentation-outdoor}).

Training PointNet++ without data augmentation provides substantially better results as it achieves a 99.1\% of precision and a recall of 98.6\%. Specifically, in the baseline PointNet++ detects 112 pedestrians out of the 345 present in the test dataset, and by not using data augmentation, this number increases to 340. For this reason, we can affirm that although applying data augmentation to the training dataset is usually beneficial in many deep learning applications~\cite{dataaugmentation}, in our case it has a negative impact on the results.

We have verified that PointNet++ can learn to detect pedestrians from 3D geometric information of point clouds in an outdoor scene. Even so, we experiment a little more to see if the performance of the neural networks improves by changing some other factors, in addition to the data augmentation. From now on, we will compare the results obtained with those of this second experiment, which we consider the {\bf new baseline}.

\subsubsection{Non-pedestrian 2D bounding boxes with less overlap}

As detailed in Section~\ref{sec:bbnp}, the average overlap between non-pedestrian 2D bounding boxes (NPBB) is around 70\%. This test aims to check whether the fact of using non-pedestrian clusters so similar to each other adversely affects the performance of PointNet++. To do so, we generate the NPBB imposing a maximum overlap between them of 20\%. This restriction implies generating fewer NPBB: 41,860, so the number of pedestrian bounding boxes (PBB) is also decreased from 9,010 to 3,234 to maintain the same ratio. 

In Table~\ref{tab:outdoor-experiments} we can see how both the precision and the recall worsen with respect to the new baseline. Therefore, training PointNet++ with non-pedestrian clusters very similar to each other (with a great number of points in common) does not impair the learning of the neural network.

\subsubsection{PointNet++ architecture}

In the last experiment we train PointNet++ using an alternative model: Multi Scale Grouping (MSG). Qi \etal designed this model to deal with more realistic situations such as those of our dataset, in which the distribution of the points is not uniform throughout the point cloud.

Using the MSG model we obtain a minimal improvement in precision (99.4\%) and a subtle worsening in the recall down to 92.5\%. Consequently, it seems unnecessary to use the MSG model to reinforce PointNet++ learning, since the non-uniformity of our point clouds is well handled by the basic model.

In this way, we can close the experimentation phase by affirming that, in a very specific context and from the experiments we have carried out with Beamagine datasets, PointNet++ provides the best performance with a batch size of 32 and without applying data augmentation for the training phase.

\subsection{Comparison with state-of-the-art approaches on different datasets}

\begin{table}
\begin{center}
\begin{tabular}{lccc}
\toprule
\textbf{Method} & 
\textbf{Dataset} & 
\textbf{\# Clusters} &
\textbf{F1-score} \\
\midrule\midrule
Tang \etal~\cite{tang} & 
KITTI & 
1,799 &
70.94 \\
\midrule
Lin \etal~\cite{lin} & 
KITTI &
1,799 &
80.11 \\
\midrule\midrule
Ours & 
Beamagine &
3,385 &
98.8 \\
\bottomrule
\end{tabular}
\end{center}
\caption{Comparison between our method and prior works on a different dataset.}
\label{tab:results-comparison}
\end{table}

We present our results using datasets obtained with Beamagine's L3CAM sensor, which incorporates an RGB camera and a high-resolution LIDAR. This allows capturing scenes with high-quality images and dense point clouds, and (partly) thanks to this we have obtained very good results. 

In order to allow a fair comparison with state-of-the-art methods and reproducible research, we attach detailed information of the Beamagine datasets in the supplementary material.

Tang \etal~\cite{tang} and Lin \etal~\cite{lin} present their results using the KITTI dataset. Although Beamagine datasets present different properties (for example in terms of vertical resolution), a comparison between these methods and the one proposed in our paper (Table~\ref{tab:results-comparison}) gives a rough view of how does our approach detecting pedestrians using PointNet++ compare with them. We have tested our method on 3,385 3D clusters, whereas KITTI contains 1,799 test clusters.

\section{Conclusions}

In this paper, we present a system to detect pedestrians in point clouds from the 3D geometric information of the scene using PointNet++ as the classifier. This neural network is designed to classify CAD models like those of the Modelnet40 dataset, making it a challenge to retrain it for classification in outdoor scenes obtained using a LIDAR sensor.

To train PointNet++ in a supervised manner, we need labelled pedestrian and non-pedestrian 3D clusters. To this end, we present a semi-automatic system to label point clouds transferring 2D bounding boxes onto the 3D domain. First, we use YOLO to detect pedestrian 2D bounding boxes in our RGB images. From YOLO predictions, we generate non-pedestrian 2D bounding boxes in the same images. These labels are transferred onto the corresponding 3D point clouds using projection matrices.

The results prove that PointNet++ can learn to classify pedestrians from non-pedestrians in outdoor scenarios only from the 3D geometric information of the point clouds obtained with a LIDAR. Likewise, these results show that the labeling system has been designed properly, since PointNet++ has learned to detect pedestrians from the generated labels. Consequently, we have designed an effective and efficient semi-automatic labeling system that takes advantage of the detection technology in RGB images to generate a ground-truth in point clouds, which can be used for datasets such as those obtained with Beamagine's L3CAM sensor.

As future work, let us mention that the developed system uses PointNet++ to classify a 3D cluster as pedestrian or non-pedestrian, but we are already working to design a strategy to detect pedestrians in an entire point cloud captured with a LIDAR to take advantage of this classification. A possible approach would be to scan a point cloud of a scene and generate random clusters with a statistic similar to the 2D pedestrian bounding boxes in terms of height, width and ratio, and input these clusters to PointNet++ to discern which are pedestrians and which are not.

It should not be forgotten that the purpose of this paper is to explore the potential contribution of 3D geometric information alone in pedestrian detection systems. This does not imply leaving aside the RGB information of the images. It would be naive not to combine both types of data to obtain a more reliable and robust detector than one that depends solely on geometric or photometric information. We may always combine both types of sensors (RGB cameras and LIDARs) to get improved results either via early or late fusion approaches.

{\small
\bibliographystyle{unsrt}
\bibliography{egbib}
}

\end{document}